\documentclass[submission,copyright,creativecommons]{eptcs}
\providecommand{\event}{SOS 2007} 
\usepackage{breakurl}             
\usepackage{underscore}           
\usepackage{amsfonts}
\usepackage{amsmath}
\usepackage{booktabs}
\usepackage[ruled,linesnumbered]{algorithm2e}
\usepackage{graphicx}
\usepackage[dvipsnames]{xcolor}

\newcommand{\ie}{\textit{i}.\textit{e}. }
\newcommand{\eg}{\textit{e}.\textit{g}. }

\title{Google vs IBM: A Constraint Solving Challenge\\ on the Job-Shop Scheduling Problem}
\author{Giacomo Da Col and Erich C. Teppan
\institute{Universitaet Klagenfurt, Austria}
}

\def\titlerunning{Google vs IBM: A Constraint Solving Challenge on the Job-Shop Scheduling Problem}
\def\authorrunning{Giacomo Da Col and Erich C. Teppan}

\begin{document}
\maketitle

\begin{abstract}
The job-shop scheduling is one of the most studied optimization problems from the dawn of computer era to the present day. Its combinatorial nature makes it easily expressible as a constraint satisfaction problem. In this paper, we compare the performance of two constraint solvers on the job-shop scheduling problem. The solvers in question are: OR-Tools, an open-source solver developed by Google and winner of the last MiniZinc Challenge, and CP Optimizer, a proprietary IBM constraint solver targeted at industrial scheduling problems. The comparison is based on the goodness of the solutions found and the time required to solve the problem instances. First, we target the classic benchmarks from the literature, then we carry out the comparison on a benchmark that was created with known optimal solutions, with sizes comparable to real-world industrial problems.
\end{abstract}

\section{Introduction}

Industrial scheduling has been one of the most investigated combinatorial problems since the Sixties\cite{fisher}. Since then, many formal definitions of such problem have been given (\eg job-shop, open-shop, flow-shop), in order to extrapolate the core aspects of the problem and neglect the insignificant ones.  

The job-shop scheduling problem\cite{adams} gained particular fame due to its easy formulation leading to instances hard to solve optimally. The most typical optimization criteria is the minimization of the makespan, \ie the time interval between the start of the first operation and the end of the last. The problem is presented as a set of jobs that must be processed by a set of machines. Each job is a sequence of operations, each operation has to be processed by a specific machine and takes a certain processing time. Every job has a specific order of operations that must be respected. An admissible solution for this problem is a sequence of operations on every machine where there is no time overlap between two operations in the same machine and the orders of the operations are respected.

Due to its combinatorial structure, it comes natural to represent this problem as a constraint satisfaction problem. In fact, constraint-based approaches have been applied successfully to job-shop problems over the past years\cite{apple,fox1982job,Sadeh96variableand}. 

A more recent technique is Large Neighboorhood Search (LNS)\cite{laborie2007self}, which consists in a continuous relaxation and re-optimization of the problem, allowing iterative improvements of the solution. This idea was also applied to MIP approaches (in the form of Relaxation Induced Neighborhood Search \cite{danna2003integrating}). In fact, hybrid approaches with CP and MIP were proposed \cite{laborie2016}, which were used in case of non-regular objective function (like in case of earliness costs). 

Despite these advancements in constraint solving, the last decade has experienced a decrease of research interest of CP applied to job-shop. Part of the problem is that the benchmarks widely used in literature (See Section 2.2) are typically more than 20 years old and are not up to date with the current industrial demands. In fact, nowadays industry can easily require up to 2000 jobs to be scheduled on 100 machines\cite{dacol}. In comparison, the biggest instance of the Taillard benchmark\cite{taillard}, which reflected real dimensions of industrial problems in 1993 and it is still among the largest available, has 50 jobs on 20 machines.

The de-facto leader on the scheduling scene of the last years is IBM, with their proprietary solver CP Optimizer. This solver was capable of finding better solutions for many job-shop instances from the classic benchmarks\cite{vilim2015failure}, as well as targeting industrial-size instances from the IBM scheduling benchmark, with instances up to 1 million activities\cite{laborie2018ibm}. However, these instances are not publicly available.  

In this paper we investigate the capabilities of the best available CP solvers on both classic benchmarks from the literature as well as on industrial-size instances with proven optima\cite{teppanbench}. By doing so, we aim to close the gap on the job-shop research of the last years.

As anticipated, one of the most successful CP solvers on the scheduling problems is CP Optimizer (abbreviated CPO). To find a worthy opponent, we took the winner of the MiniZinc Challenge 2018\footnote{https://www.minizinc.org/challenge2018/challenge.html}. The MiniZinc challenge is a recurring competition where all the constraint solvers that support the MiniZinc modeling language compete on various combinatorial problems, including scheduling. OR-Tools\footnote{https://developers.google.com/optimization/} (ORT), an open-source solver developed by Google, won the gold medal in all categories in 2018. This paper is in the same line of research as \cite{dacolandteppancp}. However, in opposition to \cite{dacolandteppancp}, we use a large-scale benchmark with proven optima herein. Furthermore, we extend the experimental setting such that, additionally to single core experiments, we also report on experiments using four processing cores (quad core).

\section{Experimental Setup}
The goal of the experiment is to compare the solving capabilities of IBM's CP Optimizer and Google's OR-Tools in jop-shop problem instances with respect to quality of the solutions (makespan) and solving time. The solvers compete on two benchmarks: one composed by classic instances from the literature, and the other is a large-scale benchmark with known optimal solutions. 

Concerning the classic benchmark, the comparison follows the rules of the MiniZinc challenge; solvers are given 20 minutes per problem instance. Concerning the large-scale benchmark we give 6 hours to complete the search. In fact, our aim is to simulate and industrial scenario, where the calculation of the schedule for the day is typically done overnight. We test the performance of the solvers with, both, a single core configuration and a quad core configuration.
 
Concerning the solvers' version, we use version 12.8.0 for CP Optimizer and version 6.10.6025 for OR-Tools. In CPO we selected the default search parameters, which turned out to be the most effective after a preliminary test. In ORT we decided to use the CP-SAT solver, because the old CP solver is not updated any more by the Google researchers, and because CP-SAT proved to be better on average after a pre-test. 
The experiment is conducted on a system equipped with a 2 GHz AMD EPYC 7551P 32 Cores CPU and 256 GB of RAM.
\subsection{Models}
There are various ways to model the job-shop problem. MiniZinc, one of the most famous CP modeling languages, is supported by ORT but it is not its native modeling language, while OPL is the native modeling language for CPO as it does not require further translation. However, both programs offer Java APIs to interface with the CP solver. To avoid bias and to make the solvers' comparison as fair as possible, we used Java to model the problem for both cases, using the same constructs and constraints. In fact, both models take advantage of the Interval Variables, a problem specific variable type that is well suited to represent job operations, because it automatically ensures that for each operation, $end = start + len$. Each machine contains a no-overlap constraint, which roughly corresponds to a cumulative constraint with the capacity set to 1. The following snippet shows the pseudo-code for the model implementations\footnote{complete encodings and benchmarks are available at https://goo.gl/qarP3m}:
 
\begin{algorithm}[H]
\small
\KwData{opDurations : IntegerArray[1..numJobs][1..numMachines]\\ \ \ \ \ \ \ \ \ \ \ opSuccessors : IntegerArray[1..numJobs][1..numMachines]}
 ops : IntervalVariableArray[1..numJobs][1..numMachines]\\
 \For{$j := 1$ \KwTo $numJobs$}{
 \For{$m := 1$ \KwTo $numMachines$}{
 		impose ops[j][m].end $\leq$ ops[j][opSuccessors[j][m]].start
 		}}
\For{$m := 1$ \KwTo $numMachines$}{
 	impose noOverlap(ops[*][m])
 		}
 minimize max(\{op.end : op $\in$ ops\})

 \caption{Job-shop encoding}
 \label{alg:ad}
\end{algorithm}

\subsection{Problem Instances}
Our test for the models are conducted on the classic benchmark and the large-scale benchmark. All the instances of the classic benchmark are rectangular job-shop instances. This means that every job has to go through all the machines, therefore every job will have a number of operations equal to the total number of machines and every machine will have assigned a number of operations equal to the total number of jobs.
The classic benchmark\footnote{https://github.com/MiniZinc/minizinc-benchmarks} consists of 74 problem instances selected from the most used job-shop benchmarks in the literature:
\begin{itemize}
\item \textbf{FT:} This benchmark is one of the oldest for job-shop scheduling, and is defined in the book "Industrial Scheduling"\cite{fisher}. It includes 3 problem instances of sizes 6x6, 10x10 and 20x5. The square instance 10x10 is famous for remaining unsolved for more than 20 years.
\item \textbf{LA:} This benchmark contains 40 problem instances from 10x5 to 30x10 \cite{lawrence}.
\item \textbf{ABZ:} 5 problem instances from the work about shifting bottleneck by \cite{adams}.
\item \textbf{ORB:} 10 problem instances proposed by \cite{apple}.
\item \textbf{YN:} 1 randomly generated problem instances of size 20x20 \cite{yamada}.
\item \textbf{SWV:} A set of 14 problem instances from \cite{storer}.
\item \textbf{VW:} 1 instance from \cite{vazquez}.
\end{itemize}

The instances of the large-scale benchmark are 24 instances divided in 8 groups of 3 by size, from 100 to 1000 machines and from 10 000 to 100 000 operations. All the instances have the optimum makespan set to 600 000 seconds, which roughly corresponds to a week. Furthermore, there are 2 types of instances:
\begin{itemize}
\item Long jobs: Less jobs with longer chain of operations;
\item Short jobs: More jobs with shorter chain of operations;
\end{itemize}

Full specification of the benchmark can be found in\cite{teppanbench}.
\begin{table}
\scalebox{0.8}{
\begin{tabular}{@{}l@{}||@{}l@{}|@{}l@{}||@{}|@{}l@{}|@{}l@{}||}
\multicolumn{1}{c}{\textbf{}} & \multicolumn{2}{c}{\textbf{single core}}  & \multicolumn{2}{c}{\textbf{quad core}}\\
\textbf{} & \textbf{CPO} & \textbf{ORT} & \textbf{CPO} & \textbf{ORT}\\
\textbf{Inst.} & \textbf{msp (secs)} & \textbf{msp (secs)} & \textbf{msp (secs)} & \textbf{msp (secs)}\\

\hline
\textbf{abz5} & 1234 (1.9) & 1234 (1.8) & 1234 (3.3) & 1234 (1.6)\\
\textbf{abz6} & 943 (0.7) & 943 (0.7) & 943 (1.4) & 943 (0.4)\\
\textbf{abz7} & 656 (1169.3) & 660 & 656 (525) & 661 (1200)\\
\textbf{abz8} & 682 & 679 & 680 & 679\\
\textbf{abz9} & 685 & 695 & 694 & 689\\
\textbf{ft06} & 55 (0) & 55 (0) & 55 (0) & 55 (0)\\
\textbf{ft10} & 930 (3.8) & 930 (5) & 930 (5.9) & 930 (2.9)\\
\textbf{ft20} & 1165 (1.4) & 1165 (5) & 1165 (0.5) & 1165 (3.4)\\
\textbf{la01} & 666 (0) & 666 (0.1) & 666 (0) & 666 (0.1)\\
\textbf{la02} & 655 (0.3) & 655 (0.1) & 655 (0.5) & 655 (0.1)\\
\textbf{la03} & 597 (0.1) & 597 (0.1) & 597 (0.1) & 597 (0)\\
\textbf{la04} & 590 (0.4) & 590 (0.2) & 590 (0.3) & 590 (0.1)\\
\textbf{la05} & 593 (0) & 593 (0) & 593 (0) & 593 (0.1)\\
\textbf{la06} & 926 (0) & 926 (1.1) & 926 (0) & 926 (0.4)\\
\textbf{la07} & 890 (0) & 890 (0.1) & 890 (0.1) & 890 (0.2)\\
\textbf{la08} & 863 (0) & 863 (0.2) & 863 (0) & 863 (0.1)\\
\textbf{la09} & 951 (0) & 951 (0.5) & 951 (0) & 951 (0.2)\\
\textbf{la10} & 958 (0) & 958 (0.9) & 958 (0) & 958 (0.1)\\
\textbf{la11} & 1222 (0) & 1222 (0.6) & 1222 (0) & 1222 (0.2)\\
\textbf{la12} & 1039 (0.1) & 1039 (0.6) & 1039 (0.2) & 1039 (0.3)\\
\textbf{la13} & 1150 (0) & 1150 (2.7) & 1150 (0) & 1150 (0.4)\\
\textbf{la14} & 1292 (0) & 1292 (1.9) & 1292 (0) & 1292 (0.3)\\
\textbf{la15} & 1207 (0.1) & 1207 (5.8) & 1207 (0.2) & 1207 (1.8)\\
\textbf{la16} & 945 (1.5) & 945 (0.6) & 945 (1.8) & 945 (0.4)\\
\textbf{la17} & 784 (1.1) & 784 (0.4) & 784 (1.4) & 784 (0.3)\\
\textbf{la18} & 848 (0.9) & 848 (1) & 848 (1.3) & 848 (0.6)\\
\textbf{la19} & 842 (2.9) & 842 (1.7) & 842 (3.4) & 842 (1.1)\\
\textbf{la20} & 902 (1.6) & 902 (0.7) & 902 (1.4) & 902 (0.6)\\
\textbf{la21} & 1046 (22.9) & 1046 (83.4) & 1046 (51.4) & 1046 (84.4)\\
\textbf{la22} & 927 (5.1) & 927 (6.2) & 927 (3.8) & 927 (4.5)\\
\textbf{la23} & 1032 (0.1) & 1032 (2.9) & 1032 (0.4) & 1032 (1.7)\\
\textbf{la24} & 935 (15.4) & 935 (24.8) & 935 (12.9) & 935 (14.2)\\
\textbf{la25} & 977 (14.5) & 977 (19.1) & 977 (19.6) & 977 (24.2)\\
\textbf{la26} & 1218 (7.4) & 1218 (79.8) & 1218 (0.8) & 1218 (11.5)\\
\textbf{la27} & 1235 (127.6) & 1235 (509.9) & 1235 (1077.9) & 1235 (479.9)\\
\textbf{la28} & 1216 (17.7) & 1216 (14.5) & 1216 (6.6) & 1216 (7.3)\\
\textbf{la29} & 1152 & 1153 & 1152 & 1152\\
\end{tabular}

\begin{tabular}{@{}l@{}||@{}l@{}|@{}l@{}||@{}|@{}l@{}|@{}l@{}|}
\multicolumn{1}{c}{\textbf{}} & \multicolumn{2}{c}{\textbf{single core}}  & \multicolumn{2}{c}{\textbf{quad core}}\\
\textbf{} & \textbf{CPO} & \textbf{ORT} & \textbf{CPO} & \textbf{ORT}\\
\textbf{Inst.} & \textbf{msp (secs)} & \textbf{msp (secs)} & \textbf{msp (secs)} & \textbf{msp (secs)}\\

\hline

\textbf{la30} & 1355 (0.3) & 1355 (21.2) & 1355 (0.7) & 1355 (8.3)\\
\textbf{la31} & 1784 (0.4) & 1784 (24.1) & 1784 (0.5) & 1784 (11.6)\\
\textbf{la32} & 1850 (0) & 1850 (29.4) & 1850 (0.1) & 1850 (20.5)\\
\textbf{la33} & 1719 (0.3) & 1719 (14.6) & 1719 (0.4) & 1719 (35.3)\\
\textbf{la34} & 1721 (1.6) & 1721 (69.5) & 1721 (1) & 1721 (31.5)\\
\textbf{la35} & 1888 (0.2) & 1888 (25.7) & 1888 (0.4) & 1888 (14.7)\\
\textbf{la36} & 1268 (10.4) & 1268 (11.3) & 1268 (6.8) & 1268 (6.6)\\
\textbf{la37} & 1397 (4) & 1397 (8.7) & 1397 (8.4) & 1397 (4.6)\\
\textbf{la38} & 1196 (85.1) & 1196 (265.1) & 1196 (108.6) & 1196 (134.7)\\
\textbf{la39} & 1233 (5.9) & 1233 (14.2) & 1233 (10.6) & 1233 (5.3)\\
\textbf{la40} & 1222 (10) & 1222 (53.2) & 1222 (31) & 1222 (30.5)\\
\textbf{orb01} & 1059 (7.1) & 1059 (22.9) & 1059 (9.6) & 1059 (19.6)\\
\textbf{orb02} & 888 (2.2) & 888 (2) & 888 (3.3) & 888 (1.2)\\
\textbf{orb03} & 1005 (6.6) & 1005 (20.9) & 1005 (17.8) & 1005 (10.7)\\
\textbf{orb04} & 1005 (2.8) & 1005 (3) & 1005 (3.8) & 1005 (1.9)\\
\textbf{orb05} & 887 (3.6) & 887 (2.4) & 887 (3.9) & 887 (2.3)\\
\textbf{orb06} & 1010 (4.7) & 1010 (8.7) & 1010 (4.7) & 1010 (6.6)\\
\textbf{orb07} & 397 (1.6) & 397 (1.4) & 397 (2) & 397 (1.1)\\
\textbf{orb08} & 899 (1.1) & 899 (1.4) & 899 (1.5) & 899 (1)\\
\textbf{orb09} & 934 (1.2) & 934 (1.5) & 934 (2) & 934 (1.1)\\
\textbf{orb10} & 944 (0.7) & 944 (1.9) & 944 (1.2) & 944 (1.1)\\
\textbf{swv01} & 1445 & 1412 & 1407 (909) & 1415\\
\textbf{swv02} & 1491 & 1475 (906.2) & 1475 (863) & 1475 (192.5)\\
\textbf{swv03} & 1420 & 1410 & 1398 (938.8) & 1415\\
\textbf{swv04} & 1520 & 1482 & 1517 & 1488\\
\textbf{swv05} & 1424 (1138.5) & 1436 & 1427 & 1430\\
\textbf{swv06} & 1728 & 1746 & 1723 & 1722\\
\textbf{swv07} & 1672 & 1677 & 1690 & 1653\\
\textbf{swv08} & 1785 & 1855 & 1872 & 1832\\
\textbf{swv09} & 1713 & 1715 & 1733 & 1694\\
\textbf{swv10} & 1823 & 1807 & 1810 & 1814\\
\textbf{swv11} & 3041 & 3317 & 3095 & 3239\\
\textbf{swv12} & 3114 & 3358 & 3056 & 3312\\
\textbf{swv13} & 3205 & 3421 & 3161 & 3321\\
\textbf{swv14} & 3032 & 3162 & 2985 & 3095\\
\textbf{vw3x3} & 256 (0) & 256 (0) & 256 (0) & 256 (0)\\
\textbf{yn4} & 980 & 994 & 993 & 994\\
  
\end{tabular}
}
\label{table1}
\caption{Results on the classic benchmark: Optimal makespans (msp) have the actual solving time in parenthesis}
\end{table}

\section{Results}
Table 1 shows the results of CP Optimizer compared to OR-Tools running on the classic benchmark, on single core and quad core configurations. Since the dataset is large, we adapted the results in two columns. In the cells we indicate the best makespans achieved before the timeout occurs. If the optimal makespan is achieved, the search stops, therefore we show the actual solving time in parenthesis. To detect whether a solution is optimal, the solver calculates a lower bound, \ie an estimate of the objective below which it is impossible to find a solution (typically a solution of the relaxed scheduling problem). When the solution found is equal to the lower bound, the solution is optimal. 

Concerning the single core, CP Optimizer was able to find a better solution than OR-Tools in 13 out of 74 problem instances (about $17.5\%$ of the instances). OR-Tools found better solutions in 6 cases, about $8\%$ of the total. CP Optimizer was faster $63.5\%$ of the time, OR-Tools $18.9\%$, and in all the other cases both solvers reached the timeout of 1200 seconds.
If we would adopt the scoring system of the MiniZinc Challenge\footnote{We used the complete scoring procedure as described in https://www.minizinc.org/challenge2018/rules2018.html}, CP Optimizer would score $53.14$ points, while OR-Tools would score $22.86$ points.
CP Optimizer solved optimally 59 problem instances ($79.7\%$), compared to 58 problem instances ($78.3\%$) of OR-Tools. In particular, OR-Tools was able to exclusively find the optimum in instance \textbf{swv02}, while CP Optimizer exclusively found the optimal solution in \textbf{swv05} and \textbf{abz7}. 

By exploiting multi core, both solvers were able to slightly improve their solutions. For example, it allowed CPO to find the optimum on the instances \textbf{swv01}, \textbf{swv02} and \textbf{swv03}, or to find the optimum on \textbf{abz7} in half of the solving time. Also ORT benefited from the additional cores, being able to find the optimum on \textbf{swv02} in a quarter of the time.

Table 2 shows the results on the large-scale benchmark with known optima. In the single core experiment, CPO is able to solve optimally 16 out of 24 instances ($66.6 \%$). In general, it was able to solve almost all the short job instances to optimality (beside 2 instances, which were still very close to optimal solution). Concerning the long job instances, all the instances with 10 000 operations reached the optimal solution, while it was never possible to hit the optimum in the cases with 100 000 operations. The hardest instances to solve were the one with long jobs, 100 machines and 100 000 operations. The worst result achieved was less than 80 \% off the optimum.

Concerning ORT in single core, it was not possible to solve any of the instances to optimality.  In two occasions, namely long-1000-10000-1 and short-1000-10000-3, it was not possible to find any admissible solution within the timeout. Beside that, the hardest instances were the long jobs with 100 000 operations, where the worst result of 206 \% off the optimum (more than 3 times the optimal makespan) was scored. In general, better performance were achieved in the short jobs instances, compared to the long ones. In fact, ORT scored on average 40 \% off the optimum on the short job instances and 154 \% off on the long ones (excluding the timeout cases).
The best result is registered on a short job instance with 100 machines and 100 000 operations, which is 10 \% off the optimum. 

Concerning the quad core test, however, things changes dramatically. ORT is able to find the optimum in 7 of the long instances, even beating CPO in two 1000-100000 instances. Some improvements were also registered on the other instances. CPO, apparently, does not benefit from the quad core as much as ORT. In fact, the solutions found were just marginally better or even worst than the single core counterparts.

\begin{table}
\centering
\scalebox{0.75}{
\begin{tabular}{l||l|l||l|l|}
\multicolumn{1}{c}{\textbf{}} & \multicolumn{2}{c}{\textbf{single core}}  & \multicolumn{2}{c}{\textbf{quad core}}\\
\textbf{Instance} & \textbf{CPO} & \textbf{ORT} & \textbf{CPO} & \textbf{ORT}\\
\textbf{type-numMachines-numOps-id} & \textbf{msp (secs)} & \textbf{msp (secs)} & \textbf{msp (secs)} & \textbf{msp (secs)}\\
\hline
\textbf{longJobs-100-10000-1}    & 600000  (8096)    & 1390577  	& 600000 (6913) &	600000 (188)
\\
\textbf{longJobs-100-10000-2}    & 600000 (10399)   & 1463638  	& 600000 (7631) &	600000 (580)
\\
\textbf{longJobs-100-10000-3}    & 600000 (10294)   & 1435995  	& 600000 (8339) &	600000 (226)
\\
\textbf{longJobs-100-100000-1}   & 1077736    & 1646792  	& 1077862  &	1642753 
\\
\textbf{longJobs-100-100000-2}   & 1066971    & 1628456  	& 1066438  &	1618288 
\\
\textbf{longJobs-100-100000-3}   & 1070306    & 1644806  	& 1070616  &	1636805 
\\
\textbf{longJobs-1000-10000-1}   & 600000 (2) 		& No Solution		 	& 600000 (3) &	No Solution	
\\
\textbf{longJobs-1000-10000-2}  & 600000 (1) 		& 1162719 (4260) 	& 600000 (3) &	600000 (7)
\\
\textbf{longJobs-1000-10000-3}   & 600000 (2) 		& 1081297  	& 600000 (2) &	600000 (2)
\\
\textbf{longJobs-1000-100000-1}  & 807297    & 1722413  	& 749737  &	600000 (563)
\\
\textbf{longJobs-1000-100000-2}  & 818596    & 1838357  	& 817481  &	600000 (3002)
\\
\textbf{longJobs-1000-100000-3}  & 837938    & 1736608  	& 839195  &	1738491 
\\
\hline
\textbf{shortJobs-100-10000-1}   & 600000 (12)      & 788640  	& 600000 (17) &	762347 
\\
\textbf{shortJobs-100-10000-2}   & 600000 (12)      & 739425  	& 600000 (24) &	741028 
\\
\textbf{shortJobs-100-10000-3}   & 600000 (19)      & 752895  	& 600000 (29) &	735739 
\\   
\textbf{shortJobs-100-100000-1}  & 600000 (4384)    & 652436  	& 600000 (5281) &	650084 (5389)
\\
\textbf{shortJobs-100-100000-2}  & 600000 (4377)    & 650084  	& 600000 (5227) &	No Solution	
\\
\textbf{shortJobs-100-100000-3}  & 600000 (4287)    & 661374  	& 600000 (5435) &	6611374 (6467)
\\
\textbf{shortJobs-1000-10000-1}  & 600000 (20026)   & 1405776  	& 600000 (19865) &	1068441 
\\
\textbf{shortJobs-1000-10000-2}  & 600000 (16538)   & 1103354  	& 600000 (17375) &	1027733 
\\
\textbf{shortJobs-1000-10000-3}  & 603447    & No Solution	 			& 600699  &	No Solution	
\\
\textbf{shortJobs-1000-100000-1} & 600000 (20956)   & 822552  	& 600147  &	790129 
\\
\textbf{shortJobs-1000-100000-2} & 600000 (16094)   & 795075  	& 600057  &	791255 
\\
\textbf{shortJobs-1000-100000-3} & 600106    & 808808  	& 600142  &	805050 
\\
\end{tabular}
}
\label{table2}
\caption{Results on the large-scale benchmark: Optimal makespan(msp) = 600000.}
\end{table}

\section{Conclusion}
CPO proved to perform better in general on the classic benchmark and especially on the large-scale benchmark. ORT benefited more than CPO from the quad core configuration, which allowed ORT to find optimal solutions even in the large-scale benchmark.

To explain the difference in performance, we analized the differences of the two solvers. While both use interval variables to express the job operations, CPO uses basic types to encode the intervals, while ORT uses three variables for a single interval, slowing down the constraint propagation. The two solvers use a similar search strategy based on large neighbourhood search (LNS), which consists in an iterative relaxation and re-optimization of the scheduling problem. However, while CPO uses portfolio strategies in combination with machine learning to converge to the best neighbourhoods, ORT uses a much more simplistic approach based on random variables/constraint selection. 

Furthermore, CPO uses a ``plan B" strategy called failure directed search (FDS), which is triggered when the LNS is not able to improve the current solution. However, we tested the impact of FDS by re-running the experiment and switching off FDS, and we found the impact to be limited on the classic benchmark (some instances improved, some worsen, many were the same) and not existing on the big instances (some actually slightly improved without FDS).  

Concluding, CP Optimizer performed slightly better than OR-Tools on the classic benchmark, but was absolutely superior on the large-scale one. In fact, CP Optimizer was able to optimally solve 66\% of the large-scale instances, against 29\% of OR-Tools (quad core). By exploiting multi cores, OR-Tools was also able to find optimal solutions for the large-scale instances, showing that nowadays CP solvers in general could be successfully applied to real-world industrial problems.
\bibliographystyle{eptcs}
\bibliography{paper}
\end{document}